\documentclass[lettersize,journal]{IEEEtran}
\usepackage{amsmath,amsfonts}
\usepackage{url}
\usepackage{adjustbox}
\usepackage{algorithm}
\usepackage{algpseudocode}
\usepackage{amsmath}
\usepackage{pifont}
\usepackage{booktabs}
\usepackage{multirow}
\usepackage{tabularx}
\usepackage{float}
\usepackage{makecell}
\newcommand{\cmark}{\ding{51}}
\newcommand{\xmark}{\ding{55}}
\usepackage{cite}
\usepackage{capt-of}
\usepackage{url}

\begin{document}

\title{Bi-HIL: Bilateral Control-Based Multimodal Hierarchical Imitation Learning via Subtask-Level Progress Rate and Keyframe Memory for Long-Horizon Contact-Rich Robotic Manipulation}
\author{
Thanpimon Buamanee$^{\dag 1}$, Masato Kobayashi$^{\dag 1,2*}$, Yuki Uranishi$^{1}$
\thanks{
$^{\dag}$ Equal Contribution, 
$^{1}$ The University of Osaka, 
$^{2}$ Kobe University,\\
$^*$ corresponding author: kobayashi.masato.cmc@osaka-u.ac.jp
}
}

\maketitle

\begin{abstract}
Long-horizon contact-rich robotic manipulation remains challenging due to partial observability and unstable subtask transitions under contact uncertainty. While hierarchical architectures improve temporal reasoning and bilateral imitation learning enables force-aware control, existing approaches often rely on flat policies that struggle with long-horizon coordination.
We propose Bi-HIL, a bilateral control-based multimodal hierarchical imitation learning framework for long-horizon manipulation. Bi-HIL stabilizes hierarchical coordination by integrating keyframe memory with subtask-level progress rate that models phase progression within the active subtask and conditions both high- and low-level policies.
We evaluate Bi-HIL on unimanual and bimanual real-robot tasks, demonstrating consistent improvements over flat and ablated variants. The results highlight the importance of explicitly modeling subtask progression together with force-aware control for robust long-horizon manipulation.
For additional material, please check: \url{https://mertcookimg.github.io/bi-hil}
\end{abstract}

\section{Introduction}

Driven by population aging and labor shortages, robots are increasingly required to perform long-horizon, contact-rich manipulation in real-world environments. Such tasks unfold as sequences of subtasks, where small execution errors accumulate over time. Robust execution therefore requires both reliable task-level reasoning over extended horizons and stable low-level control under contact uncertainty~\cite{Oliver2021SurveyRobotLearningManipulation}.

Two fundamental challenges arise in this setting. 
First, long-horizon manipulation suffers from \emph{partial observability}: from limited observations, it is often unclear whether a subtask has completed or how far it has progressed. As a result, high-level policies may repeat or prematurely switch subtasks, leading to cascading failures~\cite{chen2025sarmstageawarerewardmodeling,chen2026toprewardtokenprobabilitieshidden,ma2024visionlanguagemodelsincontext}. 
Second, contact-rich execution requires force-aware control. Small perturbations such as slips or misalignments can corrupt the perceived task state, further increasing ambiguity at subtask boundaries and destabilizing hierarchical coordination~\cite{tsuji2025surveyimitationlearningcontactrich}.

For long-horizon manipulation, hierarchical frameworks primarily address the temporal reasoning challenge by decomposing behavior into high-level planning and low-level control~\cite{Sridhar2025memER,Shi2024yellAtYourRobot,Ayalew2024progressor,li2025hamsterhierarchicalactionmodels}. While keyframe-based memory (e.g., MemER~\cite{Sridhar2025memER}) and task-progress estimation improve temporal abstraction, high-level decisions are still often inferred mainly from current observations without explicitly modeling subtask-level phase progression. Consequently, transition timing remains ambiguous and hierarchical coordination unstable in practice.

To achieve robust contact-rich manipulation in real-world settings, policies must be learned from demonstrations that capture both motion and interaction forces. Imitation learning (IL) provides a practical paradigm for acquiring such policies from human demonstrations. Unilateral teleoperation records kinematic behavior but omits force information, limiting robustness under contact uncertainty~\cite{wu2024gello,zhao2023learning,fu2024mobilealohalearningbimanual,aloha2team2024aloha2enhancedlowcost}.
In contrast, bilateral control records both position and force information, enabling force-aware policy learning~\cite{adachi2018imitation,kobayashi2025alpha}.

However, even with force-rich demonstrations, many existing bilateral control-based imitation learning (Bi-IL) approaches adopt a single flat policy that implicitly infers long-term task progression from current observations~\cite{sakaino2022variablespeed,fujimoto2019timeseries,buamanee2024biactbilateralcontrolbasedimitation}. Without explicit hierarchical coordination, subtask transitions remain unstable in long-horizon settings. As a result, the temporal reasoning challenge and the force-aware control challenge are often addressed separately rather than jointly.

\begin{figure}[t]
    \centering
    \includegraphics[width =\columnwidth]{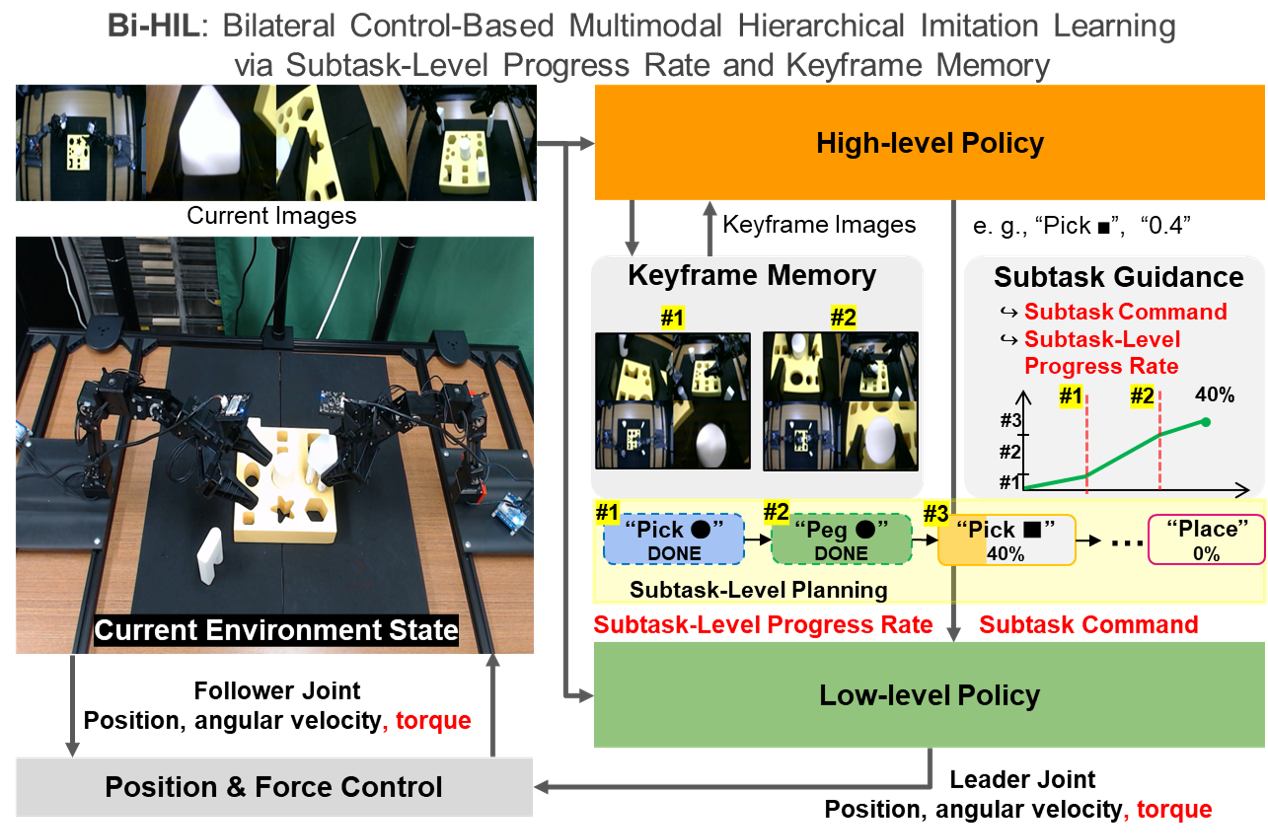}
    \caption{Concept of Bi-HIL}
    \label{TeaserFigure}
\end{figure}

To address both challenges in a unified manner, we propose Bi-HIL (Fig.~\ref{TeaserFigure}), a bilateral control-based multimodal hierarchical imitation learning framework for long-horizon, contact-rich manipulation. Bi-HIL integrates (i) keyframe memory~\cite{Sridhar2025memER} to anchor completed subtasks and (ii) subtask-level progress rate that models phase progression within the active subtask. The subtask-level progress rate is reset at each subtask transition, providing a consistent local coordination signal across hierarchical levels. Importantly, it conditions both the high-level policy and the low-level force-aware policy, enabling phase-aware short-horizon control.
\begin{table}[t]
\centering
\caption{Comparison of Long-Horizon Robot Manipulation.}
\label{tab:papers_comparison}
\resizebox{\columnwidth}{!}{
\begin{tabular}{l cc c cc cc}
\toprule
\textbf{Method}&\textbf{Hierarchy} & \textbf{Bilateral}& \textbf{Vision}& \textbf{Memory} & \textbf{Task Progress} \\
\midrule

HAMSTER~\cite{li2025hamsterhierarchicalactionmodels}
& \cmark & \xmark & \cmark 
&\xmark&\xmark\\

SARM~\cite{chen2025sarmstageawarerewardmodeling}
& \cmark  &\xmark&\cmark
&\xmark&\cmark \\

YaY Robot~\cite{Shi2024yellAtYourRobot}
&\cmark &\xmark&\cmark
&\xmark&\xmark \\

MemER~\cite{Sridhar2025memER}
&\cmark &\xmark&\cmark
&\cmark&\xmark \\

Bi-ACT~\cite{buamanee2024biactbilateralcontrolbasedimitation,yamane2025fastbilateralteleoperationimitation, kobayashi2025alpha}
&\xmark &\cmark&\cmark
&\xmark&\xmark \\

Bi-LAT~\cite{kobayashi2025bilat}, Bi-VLA~\cite{kobayashi2025bivlabilateralcontrolbasedimitation}
&\xmark&\cmark&\cmark
&\xmark&\xmark \\
Hierarchical Bi-IL~\cite{hayashi2022hierarchicalmodel}
&\cmark&\cmark&\xmark
&\xmark&\xmark \\
\textbf{Bi-HIL (Ours)}
 & \textbf{\cmark} & \textbf{\cmark} & \textbf{\cmark} 
& \textbf{\cmark} & \textbf{\cmark}  \\
\bottomrule
\end{tabular}
}
\end{table}

Our contributions are threefold:
\begin{itemize}
\item We propose Bi-HIL, a bilateral multimodal hierarchical imitation learning framework for long-horizon contact-rich manipulation.
\item We introduce a hierarchical coordination that combines keyframe memory with a resettable subtask-level progress rate, providing a subtask-local phase signal to stabilize transitions and condition low-level policy.
\item We validate Bi-HIL on real-robot unimanual and bimanual tasks, with ablations confirming that both keyframe memory and subtask progress-rate conditioning are necessary for robust contact-rich execution.
\end{itemize}

\section{Related Work}

\subsection{Framework for Long-Horizon Robotic Manipulation}

Long-horizon manipulation requires executing sequential subtasks over extended time horizons, where errors in task progression or execution can accumulate and cause failure. Hierarchical frameworks address this by decomposing behavior into high-level task planning and low-level motor control. High-level policies provide subtask commands~\cite{Shi2024yellAtYourRobot,Sridhar2025memER}, progress estimation~\cite{Ayalew2024progressor}, or motion guidance~\cite{li2025hamsterhierarchicalactionmodels} to guide low-level controller, which executes continuous control.

In imitation learning, YaY Robot predicts high-level commands from visual observations~\cite{Shi2024yellAtYourRobot}, and MemER introduces keyframe-based memory to improve temporal reasoning~\cite{Sridhar2025memER}. BPP further enhances long-context imitation by selecting semantically important history frames to reduce distribution shift~\cite{mark2026bpplongcontextrobotimitation}. Beyond architectural design, some methods explicitly encode task structure during learning: SARM employs stage-aware reward modeling~\cite{chen2025sarmstageawarerewardmodeling}, and TOPReward extracts task progress directly from internal token probabilities of video VLMs for zero-shot estimation~\cite{chen2026toprewardtokenprobabilitieshidden}.

However, as summarized in Table~\ref{tab:papers_comparison}, these approaches primarily assume unilateral control without force feedback, limiting applicability to contact-rich manipulation. Moreover, although keyframe memory provides boundary cues, it does not explicitly model subtask-level progression for low-level execution. Without progress-aware representation, transition timing remains ambiguous, leading to error accumulation.

In contrast, we propose Bi-HIL, where the high-level policy predicts subtask commands and subtask-level progress rate, while the low-level policy generates force-aware actions for robust long-horizon execution.
Unlike prior work that estimates global task completion or trajectory-level value, our method models phase progression within each individual subtask to stabilize hierarchical coordination.

\subsection{Bilateral Control-based Imitation Learning (Bi-IL)}
Bi-IL enables force-aware manipulation from demonstrations. Early work used RNN and LSTM models~\cite{adachi2018imitation,sakaino2022variablespeed,fujimoto2019timeseries}, while recent approaches adopt transformers\cite{vaswani2017attention} for improved temporal modeling and data augumentation~\cite{MasatoKobayashi2025ilbit,buamanee2024biactbilateralcontrolbasedimitation, kobayashi2024dabievaluationdataaugmentation, yamane2025fastbilateralteleoperationimitation}. Bi-ACT employs a transformer-based policy inspired by Action Chunking with Transformers (ACT)\cite{zhao2023learning} to perform force-aware tasks~\cite{kobayashi2025alpha}.
Building on this foundation, language has been further incorporated: Bi-LAT uses language instructions to modulate force magnitude during execution~\cite{kobayashi2025bilat}, while Bi-VLA uses language conditioning to disambiguate task intent under ambiguous observations~\cite{kobayashi2025bivlabilateralcontrolbasedimitation}.
As summarized in Table~\ref{tab:papers_comparison}, most Bi-IL frameworks rely on a single flat policy that implicitly infers long-horizon task progression, without explicit hierarchical coordination. This becomes problematic in long-horizon settings, where partial observability and contact-induced perturbations destabilize subtask transitions. Although Hierarchical Bi-IL~\cite{hayashi2022hierarchicalmodel} introduces hierarchical prediction of joint states, it does not leverage vision or language for global task reasoning.

In contrast, our Bi-HIL integrates hierarchical decision-making directly into Bi-IL. By introducing a high-level policy that predicts both subtask commands and a resettable subtask-level progress rate, and conditioning a force-aware low-level policy on these signals, we explicitly couple long-horizon reasoning with contact-aware control within a unified framework.

\begin{figure}[t]
    \centering
    \includegraphics[width =0.92\columnwidth]{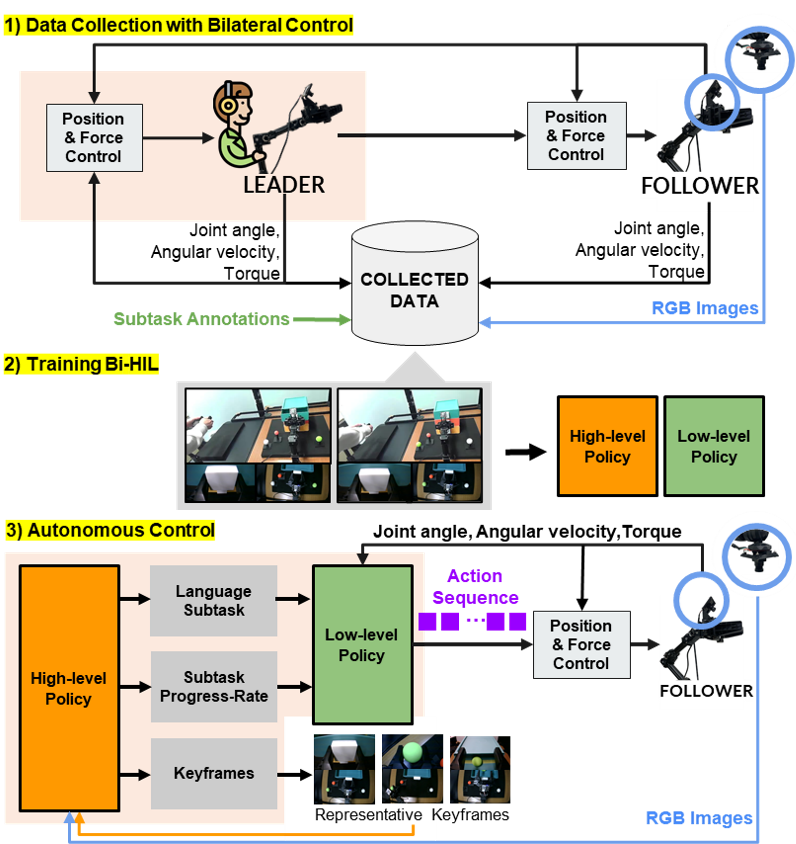}
    \caption{Overview of Bi-HIL: Bilateral Control-Based Multimodal Hierarchical Imitation Learning via Subtask-Level Progress Rate and Keyframe Memory}
    \label{overviewOfBiHIL}
\end{figure}

\section{Bi-HIL: Bilateral Control-Based Multimodal Hierarchical Imitation Learning via Subtask-Level Progress and Keyframe Memory}

\subsection{Overview}
Bi-HIL adopts a hierarchical framework consisting of a high-level policy and a low-level policy as shown in Fig.~\ref{overviewOfBiHIL}. High-level policy performs subtask-level reasoning by predicting subtask commands and subtask-level progress rate, while maintaining task memory using representative keyframes. Low-level policy predicts motor actions conditioned on visual observations, robot joint states, and high-level guidance.

During inference, the high-level policy provides temporally grounded task context, and the low-level policy executes continuous control actions based on this structured information. This hierarchical design enables robust execution of long-horizon manipulation tasks.

\subsection{Data Collection}
As shown in Fig.~\ref{overviewOfBiHIL}, Bi-HIL employs a four-channel bilateral control method for data collection, in which leader robot controlled by operator and follower robot interacts with the environment and provides force feedback to the leader. The control objective is defined as:
\begin{equation}
\theta_l - \theta_f = 0
\label{eq:position} 
\end{equation}
\begin{equation}
\tau_l + \tau_f = 0
\label{eq:force}
\end{equation}
where $\theta$ and $\tau$ denote the joint angle and torque, respectively, and the subscripts $l$ and $f$ indicate the leader and follower systems. Condition (\ref{eq:position}) enforces synchronized motion between the leader and follower, while condition (\ref{eq:force}) ensures force consistency via an action–reaction relationship. Joint angles are measured using encoders, and angular velocities are computed by numerical differentiation. Disturbance torques are estimated using a disturbance observer (DOB)~\cite{Onishi1996DOB}, and reaction torques are inferred via a reaction force observer (RFOB)~\cite{Murakami1993RFOB}.

After data collection, subtask boundaries are manually annotated based on visual observations using natural language instructions. These annotations supervise high-level policy training.
\begin{figure}[t]
    \centering
    \includegraphics[width = \columnwidth]{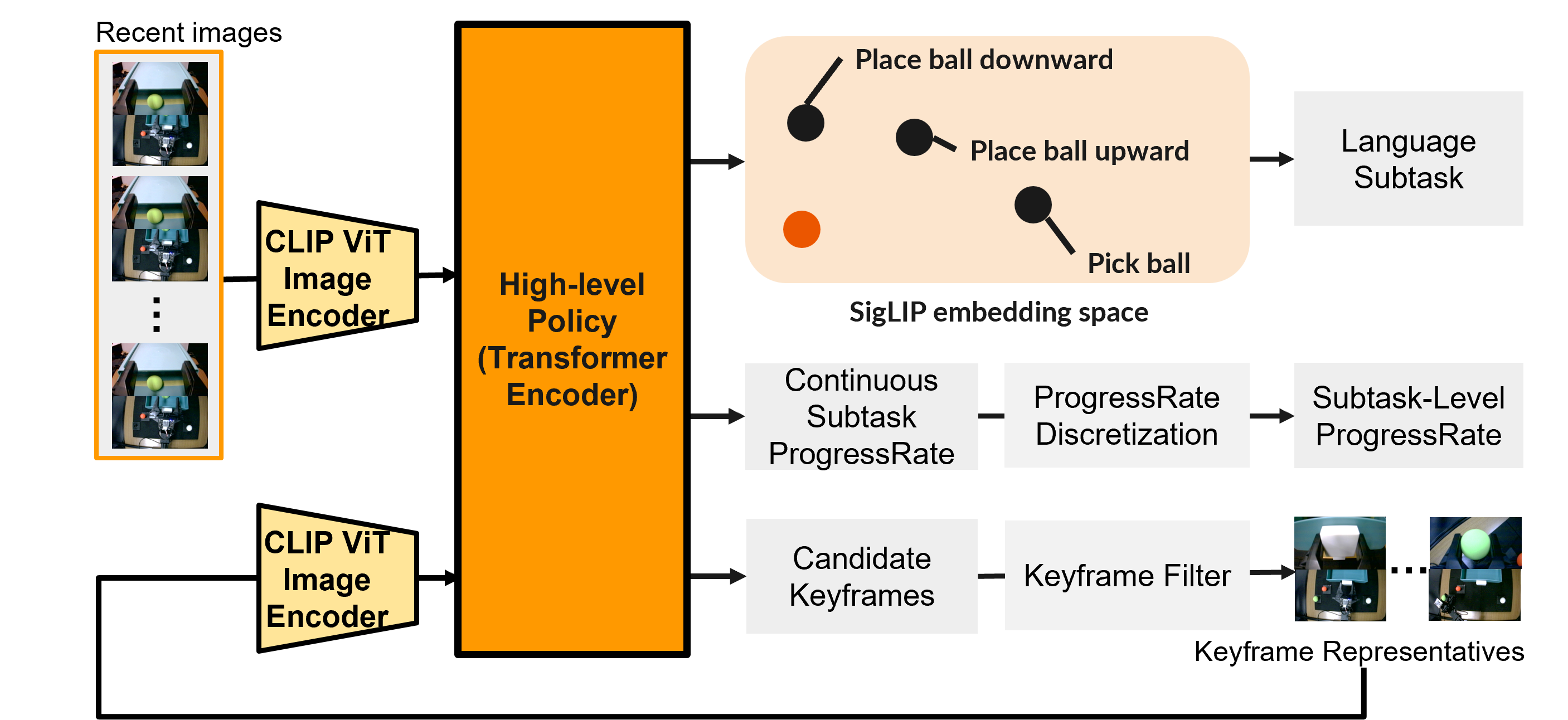}
    \caption{High-Policy Architecture}
    \label{highPolicyArchitecture}
\end{figure}

\begin{figure}[t]
    \centering
    \includegraphics[width =\columnwidth]{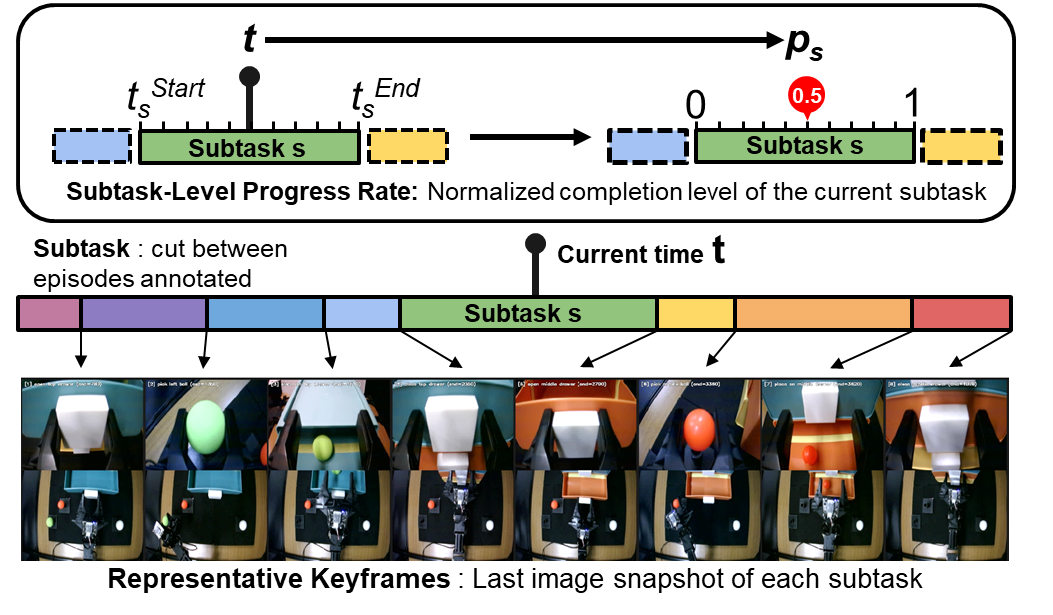}
    \caption{Definition of Representative Keyframes and Subtask-Level Progress Rate}
    \label{KeyframeProgressRateExplaination}
\end{figure}

\begin{figure}[t]
    \centering
    \includegraphics[width = \columnwidth]{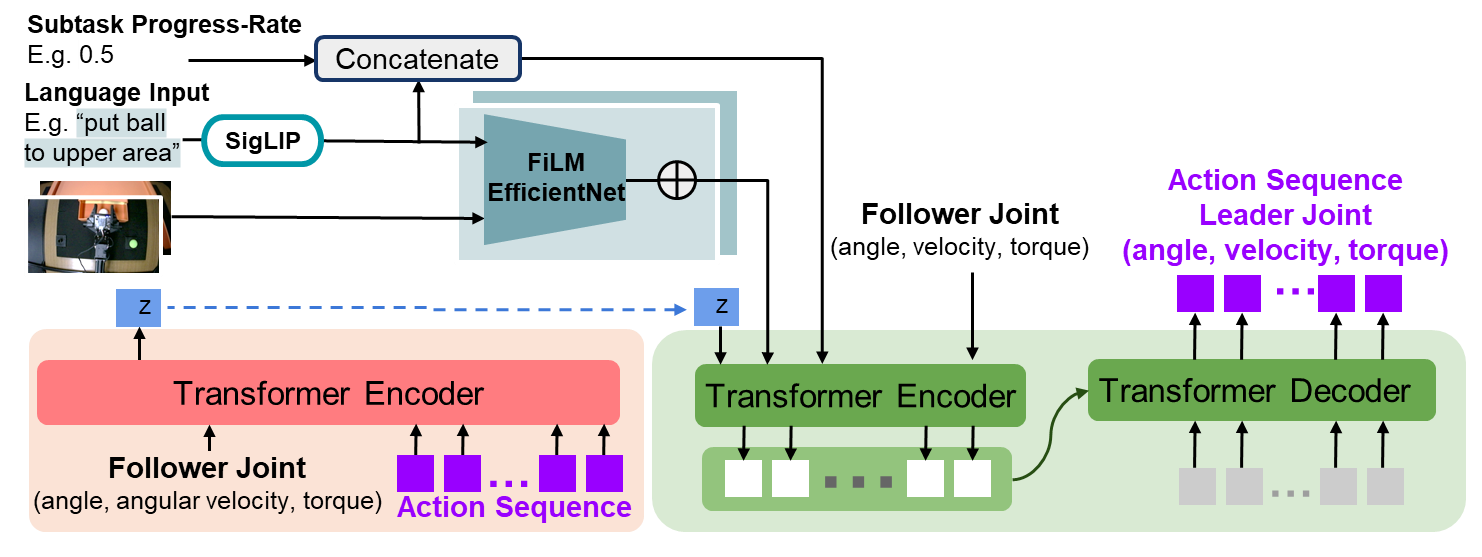}
    \caption{Low-Policy Architecture}
    \label{lowPolicyArchitecture}
\end{figure}
\begin{figure*}[t]
    \centering
    \includegraphics[width = 0.95\textwidth]{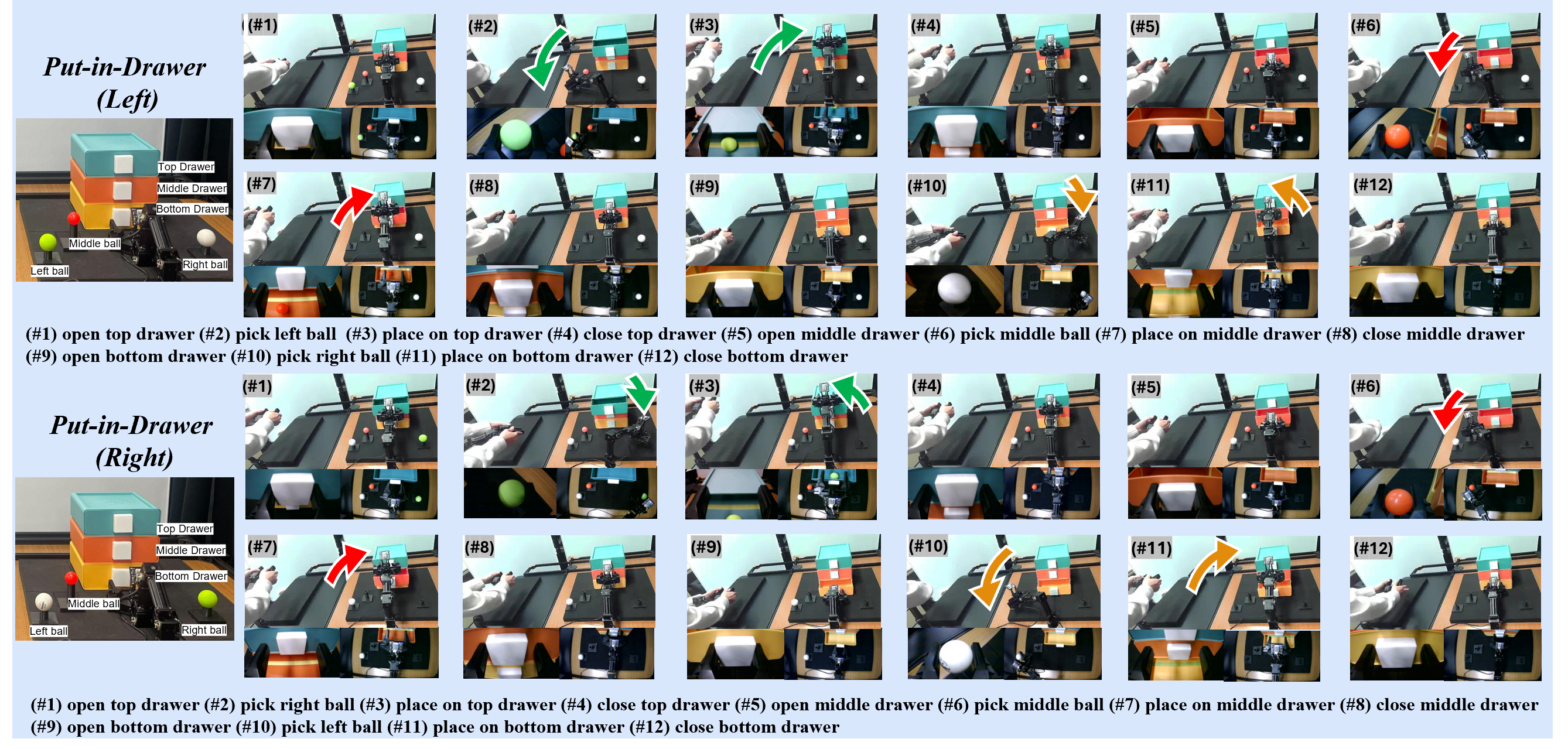}
    \caption{Data Collection of Put-Three-Balls-In-Drawer Task}
    \label{DataCollection}
\end{figure*}

\subsection{High-Level Policy}

As shown in Fig.~\ref{highPolicyArchitecture}, the high-level policy is a transformer encoder that performs subtask-level reasoning. 
It extends a YaY-style architecture~\cite{Shi2024yellAtYourRobot} with (i) keyframe-based memory inspired by MemER~\cite{Sridhar2025memER} and (ii) subtask-level progress prediction. 
These components enable temporally consistent decision-making under partial observability.

\paragraph{Inputs.}
At each high-level timestep $t$, the policy receives:
(i) a window of the most recent $N$ observations from each camera, $R_t = I_{t-N:t}$ and,  
(ii) a set of previously selected keyframes $K_t \subseteq I_{0:t-1}$, 
where $|K_t| \le K_{\max}$.
All images are encoded by a frozen CLIP encoder and processed by the transformer.

\paragraph{Outputs.}
The policy predicts three elements:
(i) a subtask command $C_t$, 
(ii) a subtask-level progress rate $p_t \in [0,1]$, and 
(iii) keyframe scores for each frame in the current window $R_t$.
Definition of representative keyframes and subtask-level progress rate are as shown in Fig.~\ref{KeyframeProgressRateExplaination}.
For a subtask $s$ with start timestep $t_s^{\mathrm{Start}}$ and end timestep $t_s^{\mathrm{End}}$, 
the subtask-level progress rate is defined as
\begin{equation}
p_s = \frac{t - t_s^{\mathrm{Start}}}{t_s^{\mathrm{End}} - t_s^{\mathrm{Start}}},
\label{eq:progressRate}
\end{equation}
which represents the normalized completion ratio of the active subtask.
The subtask-level progress rate is reset to zero when a new subtask begins.

\paragraph{Keyframe memory update.}
Positions in $R_t$ whose predicted keyframe probability exceeds a threshold $O_{th}$ are treated as candidate keyframes.
Candidate indices are accumulated over time and clustered using 1D single-linkage with a fixed temporal distance.
For each cluster, the median index is selected as the representative keyframe and stored in memory $K_t$, 
subject to the limit $K_{\max}$.
These keyframes serve as visual anchors of completed subtasks.

\paragraph{Training objective.}
The command is optimized using cross-entropy loss, the subtask-level progress rate using mean absolute error (MAE), and keyframe prediction using weighted binary cross-entropy (BCE) to address class imbalance.
The total loss is
\begin{equation}
\mathcal{L}_\mathrm{high} = \lambda_{\mathrm{cmd}}\mathcal{L}_{\mathrm{cmd}}
+ \lambda_{\mathrm{prog}}
\mathcal{L}_{\mathrm{prog}} + 
\lambda_{\mathrm{keyframe}}\mathcal{L}_{\mathrm{keyframe}},
\end{equation}
where $\lambda_{\mathrm{cmd}}$, $\lambda_{\mathrm{prog}}$, and $\lambda_{\mathrm{keyframe}}$ balances auxiliary losses.

\subsection{Low-Level Policy}
As shown in Fig.~\ref{lowPolicyArchitecture}, the low-level policy is implemented as a transformer-based conditional variational autoencoder (CVAE\cite{Sohn2015CVAE}). The model receives the subtask-level progress rate and the SigLIP-embedded subtask instruction predicted by the high-level policy, along with the current RGB images and the follower robot’s joint states (angle, angular velocity, and torque). Based on these inputs, the low-level policy predicts the next joint states of the leader robot (angle, angular velocity, and torque).
Specifically, the SigLIP-encoded subtask instruction and RGB images are first processed by a FiLM-conditioned EfficientNet, where the language embedding modulates the visual features via FiLM. Subtask-level progress rate is discretized into ten uniform levels to improve robustness. The extracted visual features and subtask-level progress rate merged with language embedding are then passed to a transformer encoder to predict the leader robot motion.
The low-level policy is trained to minimize
\begin{equation}
\mathcal{L}_{\mathrm{low}} = \mathcal{L}_{\mathrm{action}} + \beta\,\mathcal{L}_{\mathrm{KL}},
\label{eq:lowloss}
\end{equation}
where $\mathcal{L}_{\mathrm{action}}$ is the mean absolute error (L1) between the predicted and target action chunks, and $\mathcal{L}_{\mathrm{KL}}$ is the KL divergence. The weight $\beta$ balances reconstruction and regularization.

\begin{table*}[t]
  \centering
  \caption{Experimental Results: Put-Three-Balls-in-Drawer. KF: Keyframe, SPR: Subtask-Level Progress Rate.}
  \begin{tabular}{l c ccccccccccccc}
  \toprule
  \multirow{2}{*}{Model Name} & \multirow{2}{*}{Type}
  & \multicolumn{13}{c}{Completed Subtask(\%)} \\
  && \#1&\#2&\#3&\#4&\#5&\#6&\#7&\#8&\#9&\#10&\#11&\#12&Success \\
  \midrule

  \multirow{2}{*}{Bi-ACT (Baseline)}
  & Left
  &60&60&60&60&60&60&60&60&60&60&60&60
  & \multirow{2}{*}{80} \\
  & Right
  &100&100&100&100&100&100&100&100&100&100&100&100
  & \\
  \midrule

  \multirow{2}{*}{Bi-HIL (w/o KF\&SPR)}
  & Left
  &80&60&40&40&40&40&40&40&40&40&40&40
  & \multirow{2}{*}{30} \\
  & Right
  &60&20&20&20&20&20&20&20&20&20&20&20
  & \\
  \midrule

  \multirow{2}{*}{Bi-HIL  (w/o SPR)}
  & Left
  &100&20&20&20&20&20&20&20&20&20&20&20
  & \multirow{2}{*}{50} \\
  & Right
  &100&100&100&100&100&80&80&80&80&80&80&80
  & \\
  \midrule

  \multirow{2}{*}{Bi-HIL  (w/o KF)}
  & Left
  &100&100&100&100&100&100&100&100&100&60&60&60
  & \multirow{2}{*}{70} \\
  & Right
  &100&80&80&80&80&80&80&80&80&80&80&80
  & \\
  \midrule
  \multirow{2}{*}{Bi-HIL (Proposed)}
  & Left
  &100&100&100&100&100&100&100&100&100&100&100&100
  & \multirow{2}{*}{100} \\
  & Right
  &100&100&100&100&100&100&100&100&100&100&100&100
  & \\

  \bottomrule
  \end{tabular}
  \label{tab:Experimental_Results}
\end{table*}
\section{UnimanualExperiments}
\subsection{Hardware}

\begin{table}[t]
    \centering
    \caption{Model Comparison on the Put-Three-Balls-in-Drawer Task. KF: Keyframe, SPR: subtask-level Progress Rate.}
    \resizebox{\columnwidth}{!}{
    \begin{tabular}{lccccc}
        \toprule
        \multirow{2}{*}{Model Name} 
        & \multicolumn{3}{c}{High-level Output} 
        & \multicolumn{2}{c}{Low-level Input} \\\cmidrule(lr){2-4}\cmidrule(lr){5-6}
        & KF 
        & SPR
        & Command 
        & SPR
        & Command\\\toprule
        Bi-ACT (Baseline)
        & \multicolumn{3}{c}{N/A} &\xmark&\xmark\\
        Bi-HIL (w/o KF\&SPR)
        &\xmark&\xmark&\cmark&\xmark& \cmark\\
        Bi-HIL (w/o SPR)
        &\cmark&\xmark&\cmark&\xmark& \cmark\\
        Bi-HIL (w/o KF)
        &\xmark&\cmark&\cmark&\cmark& \cmark\\
        Bi-HIL (Proposed)
        &\cmark&\cmark&\cmark&\cmark& \cmark\\
        \bottomrule
    \end{tabular}
    }
    \label{tab:modelComparison}
\end{table}
As shown in Fig.~\ref{DataCollection}, unimanual experiments were conducted using OpenMANIPULATOR-X robotic arms developed by ROBOTIS. The setup consists of two robots: a leader robot operated by a human demonstrator and a follower robot interacting with the environment. Each robot is equipped with 4 degrees of freedom (DOF) for arm motion and an additional DOF for the gripper, resulting in a total of 5 actuated joints.
The control cycle was set to 1000 Hz for precise movement. Furthermore, two RGB cameras were positioned top and in the gripper area of the follower robot to record observations at 100 Hz.

\subsection{Task Setting}
We evaluate the proposed method on the long-horizon manipulation task \emph{Put-Three-Balls-in-Drawer}, in which the robot sequentially places green, red, and white balls into their corresponding drawers from top to bottom. The positions of the green and white balls are interchangeable, requiring the robot to rely on visual observations to determine the correct execution order.

This task presents three challenges: long temporal duration (59.2,s on average), visually similar subtasks, and ambiguous initial configurations. The task consists of 12 subtasks as shown in Fig.~\ref{DataCollection}, and the execution order depends on the initial ball placement. For example, when the green ball is placed on the left (Left configuration), the robot must pick the left ball first; when it is placed on the right (Right configuration), the picking order changes accordingly. This setup requires effective temporal reasoning and contextual understanding for successful execution.

\subsection{Training Setting}
Six demonstration episodes were collected using bilateral control (three Left, three Right). Each episode was 57.3--61.2 seconds. Demonstrations were manually annotated with subtask boundaries and augmented using DABI~\cite{kobayashi2024dabievaluationdataaugmentation}, increasing the dataset to 60 demonstrations. These augmented data were used to train Bi-ACT, Bi-HIL, and ablation variants of Bi-HIL, ass shown in Table~\ref{tab:modelComparison}.

We fix the following constants for reproducibility. \textbf{High-level policy:} image window length $N=5$ (history length $T=4$ plus current frame), maximum keyframes $K_{\max}=8$, and loss weights $\lambda_{\mathrm{cmd}}=\lambda_{\mathrm{prog}}=\lambda_{\mathrm{keyframe}}=1$ in Eq.~(4). The high-level policy is trained with Adam, learning rate $1\times10^{-4}$. \textbf{Low-level policy:} subtask-level progress rate is discretized into 10 levels; the transformer-based CVAE has 4 encoder and 7 decoder layers, 8 attention heads, hidden dimension 512, feedforward dimension 3200.
At inference, the high-level policy runs at $f_h=1$\,Hz and the low-level at $f_l=100$\,Hz.
The low-level policy is trained with Adam, learning rate $1\times10^{-5}$, KL weight $\beta=10$.

\begin{figure}[t]
    \centering
    \includegraphics[width = \columnwidth]{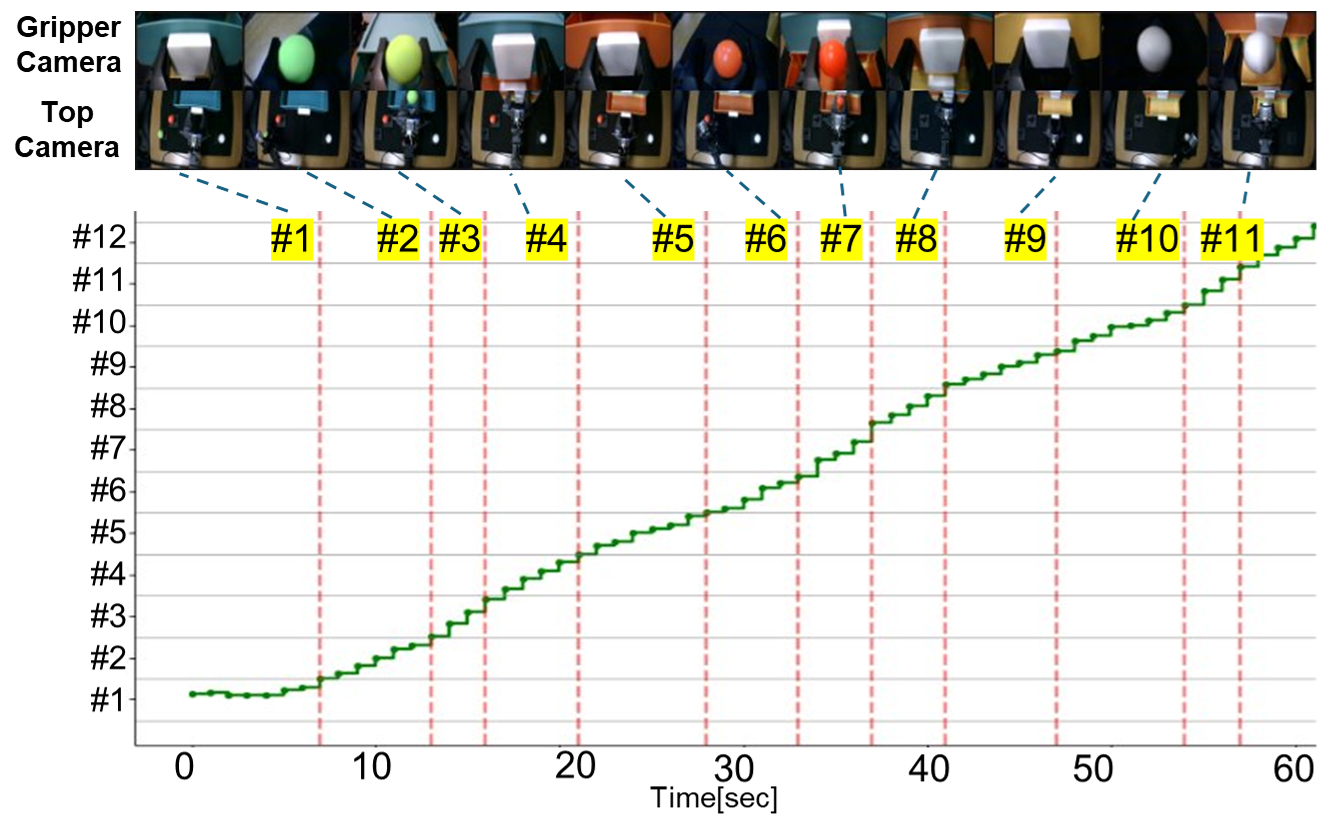}
    \caption{Put-Three-Balls-in-Drawer: Subtask-Level Progress Rate and Keyframe Memory}
    \label{HighOnly_OutputVisualization}
\end{figure}
\begin{figure*}[t]
    \centering
    \includegraphics[width =\textwidth]{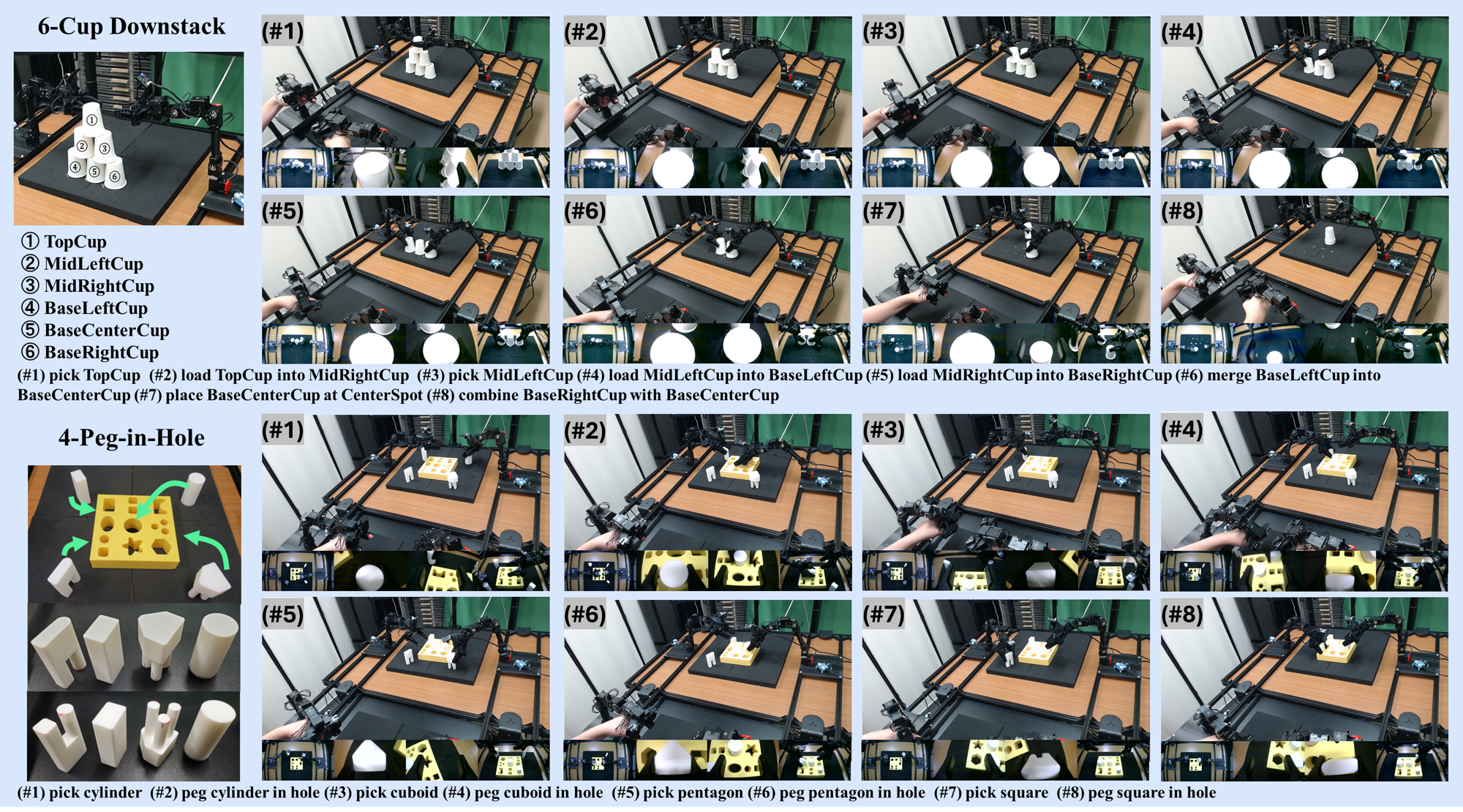}
    \caption{Data Collection of 6-Cup Downstack and 4-Peg-in-Hole}
    \label{fig:bi-tele}
\end{figure*}
\subsection{Experiment Result}
Table~\ref{tab:Experimental_Results} reports the performance of each model over ten trials: five on Put-Three-Balls-in-Drawer (Left) and five on Put-Three-Balls-in-Drawer (Right). For failed trials, the completed subtask columns indicate the furthest progress achieved by the robot before failure, ranging from subtask (\#1) Open top drawer to subtask (\#12) Close bottom drawer. The Success column reports the overall task success rate.

For the baseline method Bi-ACT, a 100\% success rate (5/5) is achieved in the Right configuration, whereas performance drops to 60\% (3/5) in the Left configuration, with the model mistakenly opening the drawer in an incorrect order. This reveals difficulty in handling ambiguous task progression and highlights the need for hierarchical reasoning.

Despite lacking a high-level policy, Bi-ACT performs second-best among the models, outperforming the Bi-HIL ablation variants that incorporate a high-level policy. This indicates that simply adding a high-level policy
does not guarantee improved performance without appropriate design.

The proposed Bi-HIL model is the only method to achieve a 100\% success rate on both configurations(5/5 for Left, 5/5 for Right), demonstrating superior performance on this task. Fig.~\ref{HighOnly_OutputVisualization} shows the evolution of predicted subtasks and subtask-level progress rate over time. The visualization shows a clear progression from lower left to upper right, indicating consistent subtask prediction and progress estimation advancement. Keyframes (red) are predicted near subtask boundaries, demonstrating effective memory selection.

Ablation results demonstrate that both keyframe memory and subtask-level progress rate are essential: commands alone lead to unstable predictions, keyframes without subtask-level progress rate lack temporal guidance, and subtask-level progress rate without keyframes lacks reliable task memory. In contrast, the full Bi-HIL model integrates both components to achieve stable reasoning and reliable execution.

\section{Bimanual Experiments}
\subsection{Hardware}
\begin{figure}[t]
    \centering
    \includegraphics[width = \columnwidth]{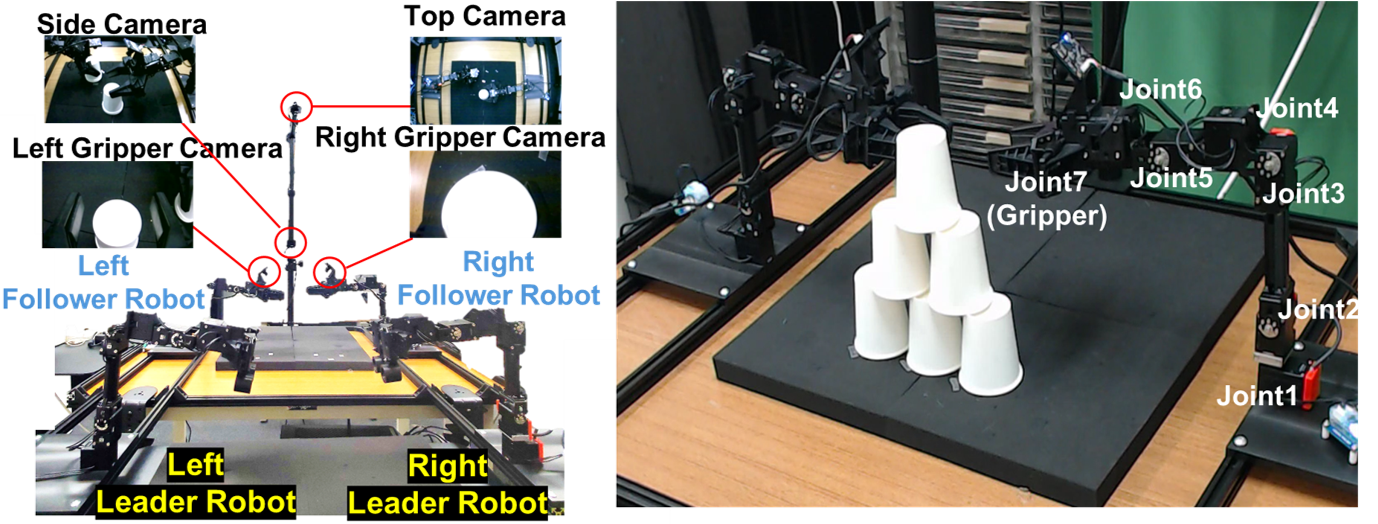}
    \caption{Experimental Setup}
    \label{fig:bi-env}
\end{figure}
As shown in Fig.~\ref{fig:bi-env}, ALPHA-$\alpha$ were used for experiments of bimanual robotic manipulation.
A total of four robots were utilized in the experiments, including two leader robots operated by the human operator and two follower robots. Each robot has six degrees of freedom (DOF) for versatile movement, as well as an additional DOF for the gripper, utilizing a total of seven motors for its operation.
The bilateral control cycle was set to 1000 Hz for precise movement and data collection of joint angle, velocity, and torque.
Furthermore, four RGB cameras were placed top, on the sides, and at both the right and left gripper areas of the follower robots to record observations.

\subsection{Task Setting}
To examine the applicability of Bi-HIL, experiments were conducted on "6-Cup Downstack" and "4-Peg-in-Hole" task, as shown in Fig.~\ref{fig:bi-tele}.

\subsubsection{6-Cup Downstack}
The 6-cup downstack is performed through a sequence of structured pick-and-insert actions, progressively nesting the upper tiers into the base layer and consolidating the cups into a single centralized stack.

\subsubsection{4-Peg-in-Hole}
The 4-Peg-in-Hole task consists of a sequence of precise pick-and-insert actions, where the robot grasps geometric pegs and inserts them into their corresponding holes. 
The 4-Peg-in-Hole task emphasizes accurate alignment and force-sensitive insertion under contact-rich conditions.
The peg and hole geometries are adopted from FMB~\cite{jianlan2025peg}. Both the pegs and holes are scaled to 0.8 times their original size and fabricated using a 3D printer.

\subsection{Training Setting}
For each task, we collected five demonstrations as training data.
Each 6-Cup Downstack episode was 45.1--47.3 seconds.
Each 4-Peg-in-Hole episode was 49.5--54.6 seconds.
Demonstrations were manually annotated with subtask boundaries and augmented using DABI~\cite{kobayashi2024dabievaluationdataaugmentation}, increasing the dataset to 50 demonstrations.
These were used to train Bi-ACT without force feedback, Bi-ACT, and Bi-HIL.
The high-level and low-level policy of parameters are the same as in the unimanual experiments.

\subsection{Experimental Results}

\subsubsection{6-Cup Downstack Evaluation}
\begin{table}[t]
  \centering
  \caption{Experimental Results: 6-Cup Downstack}
  \label{tab:6cup_results}
  \resizebox{\columnwidth}{!}{
  \begin{tabular}{l cccccccccc}
  \toprule
  \multirow{2}{*}{Model Name}
  & \multicolumn{9}{c}{Completed Subtask(\%)} \\
  &\#1&\#2&\#3&\#4&\#5&\#6&\#7&\#8&Success \\
  \midrule
  Bi-ACT (w/o Force)
  &60&60&60&60&40&40&40&20&20 \\
  Bi-ACT
  &100&100&100&100&100&80&80&60&60\\
  Bi-HIL
  &100&100&100&100&80&80&80&80&80 \\
  \bottomrule
  \end{tabular}
  }
\end{table}

\begin{figure}[t]
    \centering
    \includegraphics[width = \columnwidth]{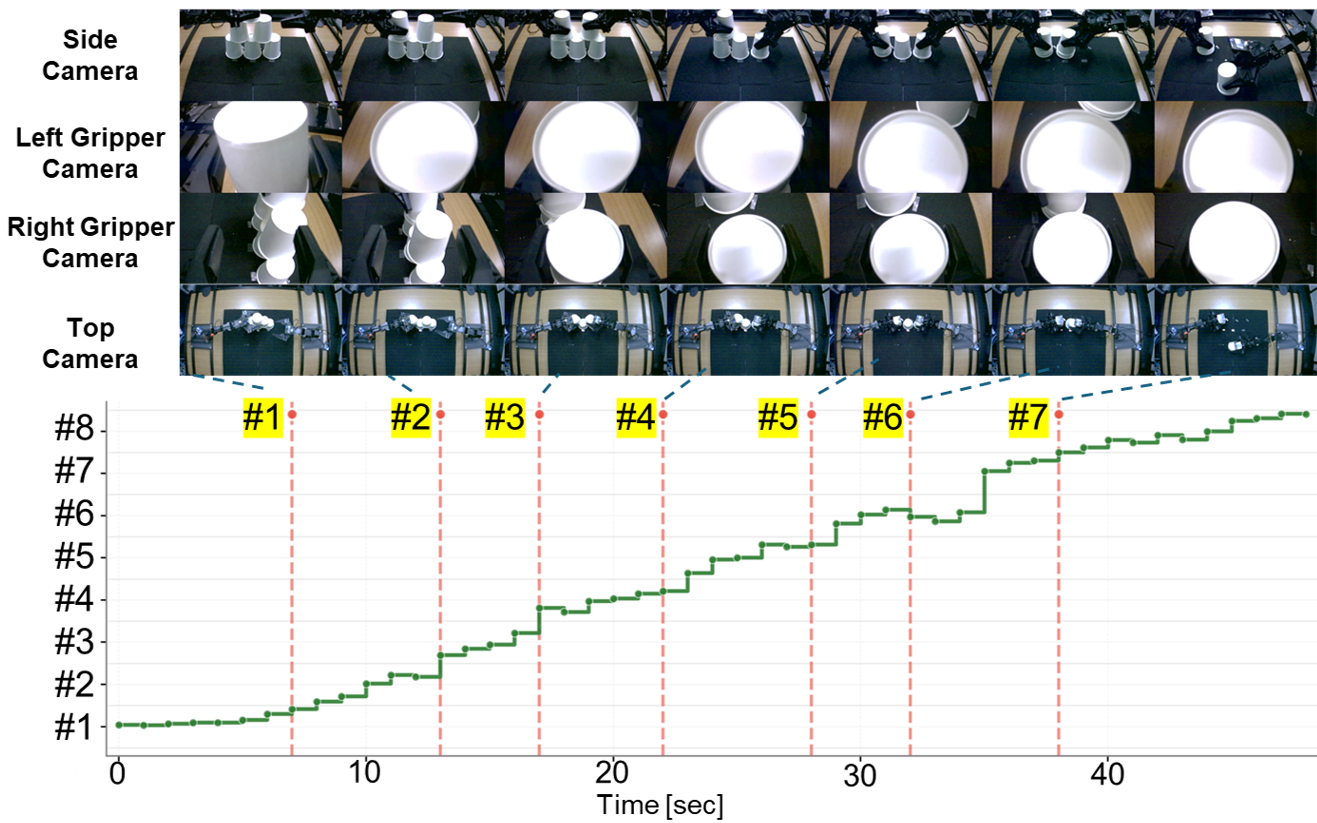}
    \caption{6-Cup Downstack: Subtask-Level Progress Rate and Keyframe}
    \label{HighOnly_OutputVisualization_bitask1}
\end{figure}
Table~\ref{tab:6cup_results} reports the experimental results on the 6-Cup Downstack task over five trials. 
For failed trials, the completed subtask indicates the furthest progress achieved before execution failure. The Success column reports the overall task success rate.

Bi-ACT without force feedback achieves only 20\% success (1/5), with performance degradation occurring after subtask 5. 
Failures were mainly caused by unstable insertion during contact-rich stacking motions, highlighting the importance of force information in bimanual manipulation.

Bi-ACT with force control improves the success rate to 60\% (3/5), successfully completing the initial stacking phases. 
However, performance decreases in later subtasks (7–8), where coordinated bimanual reasoning is required to merge the final cup structures. 
These results indicate that while force feedback improves execution stability, a flat policy without hierarchical reasoning struggles with long-horizon coordination.

The proposed Bi-HIL achieves the highest performance with an 80\% success rate (4/5). 
All trials successfully completed subtasks 1–4, and most trials progressed consistently through subtasks 5–8. 
Compared to Bi-ACT, Bi-HIL shows improved stability in the later merging stages, demonstrating that hierarchical task reasoning combined with force-aware low-level control enhances long-horizon bimanual manipulation. Fig.~\ref{HighOnly_OutputVisualization_bitask1} shows the evolution of the predicted subtasks and progress rate over time for the 6-Cup Downstack task. The visualization illustrates an overall progression from the lower left to the upper right, indicating consistent subtask prediction and progress estimation. Although slight fluctuations are observed in the subtask-level progress rate, the prediction stabilizes in subsequent steps. Keyframe candidates, shown in red, are predicted seven times, capturing the end of each subtask.

These results confirm that integrating subtask-level reasoning with bilateral control improves robustness in structured, contact-rich bimanual tasks such as the 6-Cup Downstack.

\subsubsection{4-Peg-in-Hole Evaluation}
\begin{table}[t]
  \centering
  \caption{Experimental Results: 4-Peg-in-Hole}
  \label{tab:peg_results}
  \resizebox{\columnwidth}{!}{
  \begin{tabular}{l cccccccccc}
  \toprule
  \multirow{2}{*}{Model Name}
  & \multicolumn{9}{c}{Completed Subtask(\%)} \\
  &\#1&\#2&\#3&\#4&\#5&\#6&\#7&\#8&Success \\
  \midrule
  Bi-ACT (w/o Force)
  &100&80&40&0&0&0&0&0&0 \\
  Bi-ACT
  &100&100&100&60&60&40&40&20&20 \\
  Bi-HIL
  &100&100&100&100&100&80&80&80&80 \\
  \bottomrule
  \end{tabular}
  }
\end{table}

\begin{figure}[t]
    \centering
    \includegraphics[width = \columnwidth]{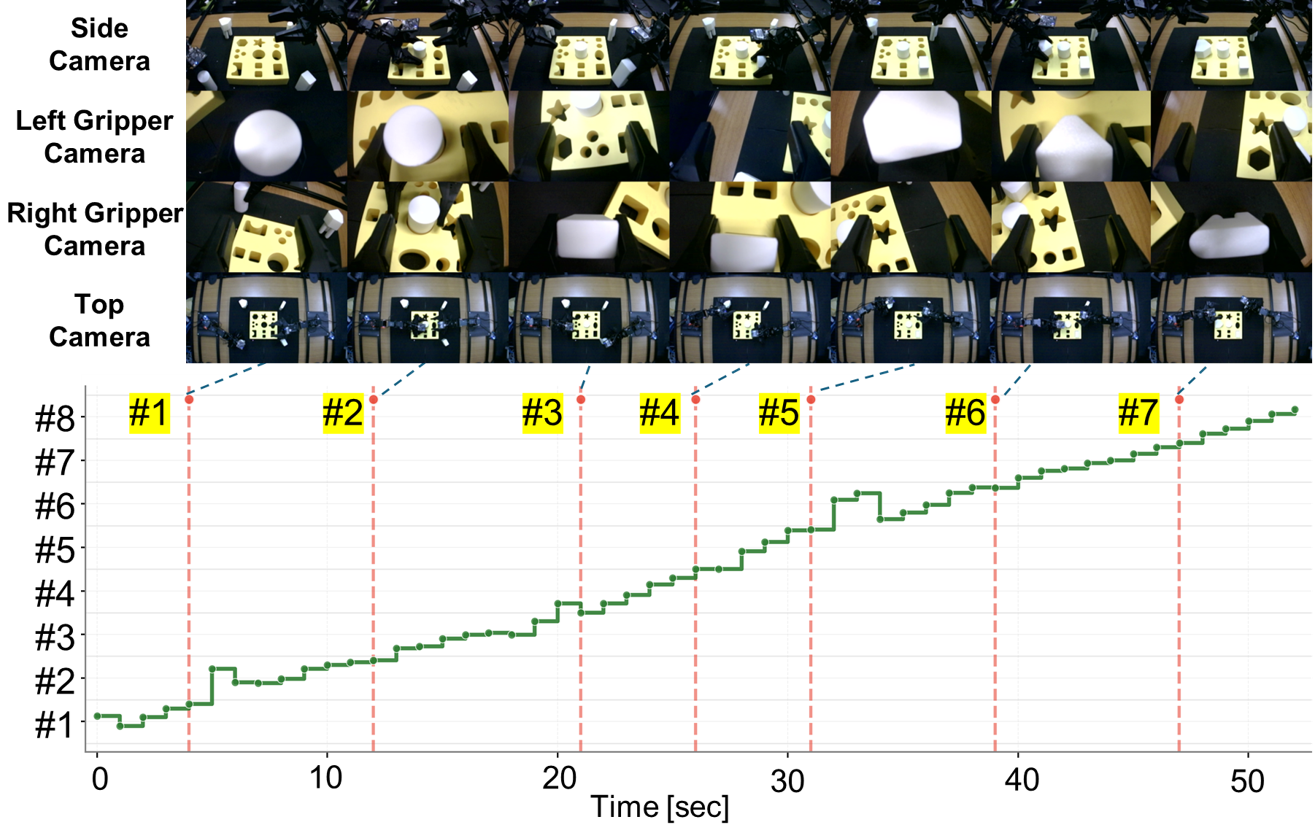}
    \caption{ 4-Peg-in-Hole: Subtask-Level Progress Rate and Keyframe}
    \label{HighOnly_OutputVisualization_bi4-Peg-in-Hole}
\end{figure}
Table~\ref{tab:peg_results} summarizes the experimental results on the 4-Peg-in-Hole task over five trials. 
For failed trials, the completed subtask indicates the furthest progress achieved prior to execution failure, and the Success column reports the overall task success rate.

Bi-ACT without force feedback fails to complete the task in all trials (0\% success). 
While the model consistently accomplishes the initial grasping subtasks (\#1–\#2), performance degrades during the first insertion phase, and no trial progresses beyond subtask \#3. 
This failure pattern indicates that vision-only control is insufficient for maintaining stable alignment and contact during peg insertion, where precise force modulation is required.

Introducing force feedback (Bi-ACT) improves early-stage execution, with all trials successfully completing subtasks \#1–\#3. 
However, the overall success rate remains limited to 20\% (1/5). 
Most failures occur in later insertion stages (\#6–\#8), where sequential coordination across multiple pegs is necessary. 
These results suggest that although force feedback enhances local contact stability, a flat policy lacks the temporal abstraction required for consistent long-horizon execution.

The proposed Bi-HIL achieves the highest performance, attaining an 80\% success rate (4/5). 
All trials successfully complete subtasks \#1–\#4, and the majority progress reliably through the remaining insertion phases. 
Compared to both Bi-ACT variants, Bi-HIL exhibits markedly improved robustness during repeated alignment and insertion operations. 
Fig.~\ref{HighOnly_OutputVisualization_bi4-Peg-in-Hole} shows that predicted subtasks and progress rate evolve smoothly over time, and keyframe candidates concentrate near subtask boundaries, indicating stable phase estimation and transition detection.

Overall, these results indicate that hierarchical task decomposition combined with force-aware low-level control substantially improves reliability in contact-rich assembly tasks. 
Explicit modeling of subtask-level progress provides structured temporal guidance that enables consistent coordination across sequential insertion phases.

\section{Conclusion}
We presented Bi-HIL, a bilateral control-based multimodal hierarchical imitation learning framework for long-horizon contact-rich manipulation. Bi-HIL couples a high-level policy that predicts subtask commands and a resettable subtask-level progress rate with a force-aware low-level policy learned from bilateral demonstrations.
Experiments on real robots show that Bi-HIL improves robustness over baselines and ablated variants on both unimanual and bimanual settings.
On the unimanual task, Bi-HIL achieves reliable long-horizon execution with stable subtask transitions. On bimanual contact-rich tasks, Bi-HIL consistently outperforms force-aware baseline policies, particularly in later subtasks. 
These results indicate that explicit subtask-level phase modeling, together with keyframe memory and force-aware control, is critical for robust long-horizon manipulation.


\vfill

\end{document}